\documentclass[conference]{IEEEtran}
\usepackage[left=0.75in,right=0.75in,top=0.75in,bottom=0.75in]{geometry}
\IEEEoverridecommandlockouts
\usepackage{graphicx} % Required for inserting images
\usepackage{cite}
\usepackage{algorithmic}
\usepackage{textcomp}
\usepackage[utf8]{inputenc} % allow utf-8 input
\usepackage[T1]{fontenc}    % use 8-bit T1 fonts
\usepackage{hyperref}       % hyperlinks
\usepackage{url}            % simple URL typesetting
\usepackage{booktabs}       % professional-quality tables
\usepackage{nicefrac}       % compact symbols for 1/2, etc.
\usepackage{microtype}      % microtypography
\usepackage{xcolor}         % colors
\usepackage{amsmath,amssymb,multicol,latexsym,amsfonts}
\usepackage{pgfplots,pgfplotstable}
\pgfplotsset{compat=1.16}
\usetikzlibrary{pgfplots.groupplots}
\usepackage{subcaption}
\usepackage{hyperref}
\usepackage{cleveref}
\usepackage{todonotes}
\usepackage{flushend} % references even

\graphicspath{{figures/}}

\def\BibTeX{{\rm B\kern-.05em{\sc i\kern-.025em b}\kern-.08em
    T\kern-.1667em\lower.7ex\hbox{E}\kern-.125emX}}

\makeatletter
\def\ps@IEEEtitlepagestyle{%
  \def\@oddfoot{\mycopyrightnotice}%
  \def\@evenfoot{}%
}
\def\mycopyrightnotice{%
  \begin{minipage}{\textwidth}
  \centering \scriptsize
  Copyright~\copyright~2025 IEEE. Personal use of this material is permitted. Permission from IEEE must be obtained for all other uses, in any current or future media, including reprinting/republishing this material for advertising or promotional purposes, creating new collective works, for resale or redistribution to servers or lists, or reuse of any copyrighted component of this work in other works.
  \end{minipage}
}
\makeatother

\author{zipfl}
\date{January 2025}
\begin{document}
\bstctlcite{IEEEexample:BSTcontrol}

\title{\vspace{0.25in}DigiT4TAF - Bridging Physical and Digital Worlds for Future Transportation Systems}

\author{
Maximilian~Zipfl$^{1}$,
Pascal~Zwick$^1$,
Patrick Schulz$^1$,
Marc René Zofka$^1$,
Albert~Schotschneider$^1$,\\
Helen Gremmelmaier$^1$,
Nikolai Polley$^{2}$,
Ferdinand Mütsch$^{2}$
Kevin~Simon$^{2}$,
Fabian~Gottselig$^{2}$,\\
Michael~Frey$^{2}$,
Sérgio Marschall$^1$,
Akim Stark$^1$,
Maximilian Müller$^1$,
Marek~Wehmer$^1$,
Mihai Kocsis$^{1,3}$,\\
Dominic Waldenmayer$^3$,
Florian Schnepf$^3$,
Erik Heinrich$^3$,
Sabrina~Pletz$^3$,
Matthias~Kölle$^{4}$,\\
Karin~Langbein-Euchner$^{4}$,
Alexander~Viehl$^{1}$,
Raoul Zöllner$^3$,
and J.~Marius~Zöllner$^{1,2}$

\thanks{$^{1}$FZI Research Center for Information Technology, Karlsruhe, Germany
{\tt\small \{surname\}@fzi.de}.}
\thanks{$^{2}$KIT - Karlsruhe Institute of Technology, Karlsruhe, Germany \hfill
{\tt\small \{prename.surname\}@kit.edu}.}
\thanks{$^{3}$ Automotive Systems Engineering, Heilbronn University of Applied Science, Heilbronn Germany \newline
{\tt\small \{prename.surname\}@hs-heilbronn.de}.}
\thanks{$^{4}$SSP Consult, Consultant Engineers, Stuttgart, Germany
{\tt\small \{surname\}@ssp-consult.de}.}
}

\maketitle

\begin{abstract}
In the future, mobility will be strongly shaped by the increasing use of digitalization. Not only will individual road users be highly interconnected, but also the road and associated infrastructure.
At that point, a Digital Twin becomes particularly appealing because, unlike a basic simulation, it offers a continuous, bilateral connection linking the real and virtual environments. 
This paper describes the digital reconstruction used to develop the Digital Twin of the Test Area Autonomous Driving—Baden-Württemberg (TAF-BW), Germany.
The TAF-BW offers a variety of different road sections, from high-traffic urban intersections and tunnels to multilane motorways. The test area is equipped with a comprehensive Vehicle-to-Everything (V2X) communication infrastructure and multiple intelligent intersections equipped with camera sensors to facilitate real-time traffic flow monitoring. The generation of authentic data as input for the Digital Twin was achieved by extracting object lists at the intersections. This process was facilitated by the combined utilization of camera images from the intelligent infrastructure and LiDAR sensors mounted on a test vehicle. 
Using a unified interface, recordings from real-world detections of traffic participants can be resimulated. Additionally, the simulation framework's design and the reconstruction process is discussed. 
The resulting framework is made publicly available for download and utilization at: \url{https://digit4taf-bw.fzi.de}\\
The demonstration uses two case studies to illustrate the application of the digital twin and its interfaces: the analysis of traffic signal systems to optimize traffic flow and the simulation of security-related scenarios in the communications sector.
\end{abstract}


% =============
% Introduction
%==============
%%%%%%%%%%%%%%%%%%%%%%%%%%%%%%%%%%%%%%%%%%
\section{Introduction}
\label{sec:introduction}
%%%%%%%%%%%%%%%%%%%%%%%%%%%%%%%%%%%%%%%%%%

%%%%%%%%%%%%%%%%%%%%%%%%%%%%%%%%%%%%%%%%%%
The increasing digitalization of transportation systems has underscored the transformative potential of Artificial Intelligence (AI) in shaping future urban mobility.

Today's urban mobility is the result of a variety of players interacting with smart, connected forms of transport and the transport infrastructure. In order to make these usable for the planning, control and optimization of various use cases from the domain of mobility, the concept of the Digital Twin, the optimization of traffic flow, the increase in traffic safety through to more efficient mobility, can be applied.

The Test Area Autonomous Driving Baden-Württemberg (TAF-BW) offers such a smart traffic infrastructure. 
Digital information in the form of high-definition maps and real-time data is collected via multisensor systems and communicated via IEEE802.11p and ITS-G5 along various reference tracks in Karlsruhe and Heilbronn, which are used to integrate connected traffic participants into the mobility system.

%%%%%%%%%%%%%%%%%%%%%%%%%%%%%%%%%%%%%%%%%%
In this context, the concept of a Digital Twin of a traffic infrastructure becomes particularly relevant. 
There are different definitions of Digital Twins. 
First, Grieves implemented the term Digital Twin with respect to Product Lifecycle Management.
Nowadays, there are definitions that take care of mobility and automotive applications, see \cite{asi5040065,PfeifferFuchssKuehn2024_1000174513,9724183,10920004}. 
Unlike a conventional simulation, which operates in isolation, a Digital Twin "is a non-physical model" \cite{asi5040065}, which establishes a continuous, bidirectional connection between physical and virtual environments. The connection between the sensors and actuators in the physical world are being integrated with simulation models, data and analytics in the virtual world with human machine interfaces to control and reason with the Digital Twin. 
This dynamic linkage allows for real-time monitoring, analysis, and adaptation, thereby enhancing the efficiency, safety, and sustainability of modern mobility systems.  
A Digital Twin not only represents actual traffic processes as a virtual image of reality, but also enables the interpretation of complex correlations and interactions between various domain models and real-time data, allowing stakeholders to perform What-If analyses by modifying external boundary conditions.

%%%%%%%%%%%%%%%%%%%%%%%%%%%%%%%%%%%%%%%%%%

This is precisely where the present paper comes in. 
The contribution of the present work is organized as follows:
In order to enable different use cases of 
\begin{itemize}
    \item vulnerable traffic participant (VRU) Safety by integrating Cyclist with means of Virtual Reality
    \item optimization of Traffic Light Signals and
    \item security issues regarding the connected mobility system.
\end{itemize}
we contribute with a harmonized approach to design the Digital Twin of the complex mobility system of the aforementioned TAF-BW. 
We emphasize how we integrate complex vulnerable traffic participant behavior with a focus on cyclists. 
The consideration of traffic lights and their implementation. 
The Digital Twin simulation framework and corresponding tooling is available for download at \url{digit4taf-bw.fzi.de}
%%%%%%%%%%%%%%%%%%%%%%%%%%%%%%%%%%%%%%%%%%

%%%%%%%%%%%%%%%%%%%%%%%%%%%%%%%%%%%%%%%%%%
Therefore, the present paper is structured as follows: 
In the following \Cref{sec:related_work} the relevant state of the art is presented. 
In \Cref{sec:digital_twin} the aforementioned contribution is explained in detail. Afterward, in \Cref{sec:experiments} we demonstrate the application of our Digital Twin regarding the different use cases 
and conclude our present work in \Cref{sec:conclusion}.

% ==============
%  State of the Art
% ==============
%%%%%%%%%%%%%%%%%%%%%%%%%%%%%%%%%%%%%%%%%%
\section{Related Work}
\label{sec:related_work}
%%%%%%%%%%%%%%%%%%%%%%%%%%%%%%%%%%%%%%%%%%

There are a number of different digital twins focusing on different aspects of mobility and transport applications.
These differ in terms of their specific area of application and granularity. For example, communication, traffic flow or individual dynamic traffic participants are simulated and information is fed back into the real world \cite{wu2025digitaltwinframeworkphysicalvirtual}.

The work Wu et al. \cite{wu2025digitaltwinframeworkphysicalvirtual} presents a simulation of a real urban area using online V2X data. The Digital Twin achieves a replication of real traffic conditions, including vehicle behaviour, signal control and road geometry. The main emphasis is on vehicles, and pedestrians are not further considered.

Wang et al. \cite{9724183} present a concept for a large-scale digital twin and its cloud and data management architecture. The focus is on the distribution of data to different stakeholders.

Mcity \cite{Liu_MCity_} is a test facility designed for early-stage research into automated and connected driving. It provides a simulation of the facility, but the area is disconnected from public roads.

The work of \cite{hossain_new_2023} gives are broad overview of applications of the Digital Twin technology and give a good overview of recent works.

The work of Klar et al. \cite{10920004} discusses recent advancements for Digital Twins in the mobility and transport sector. It provides a detailed discussion of the digital twin maturity level required to facilitate the realization of their respective objectives. The conclusion drawn is that Digital Twins are already being used to monitor critical vehicle components, and initial test beds for performing Digital-Twin-enabled autonomous vehicle operations in closed environments exist. However, it is acknowledged that several challenges must be addressed before Digital Twins can achieve their full potential. These challenges include standardization, interoperability, safety and security.

In our paper we want to address these issues and build a framework that uses standardized interfaces for V2X communication and co-simulation, allowing seamless extension to other simulation environments. We are also looking at the security of the V2X communication of our digital twin.

% ===============================
%  Structure of the Digital Twin
% ===============================
%%%%%%%%%%%%%%%%%%%%%%%%%%%%%%%%%%%%%%%%%%
\section{The Digital Twin}
\label{sec:digital_twin}
%%%%%%%%%%%%%%%%%%%%%%%%%%%%%%%%%%%%%%%%%%

\subsection{Structure of the Digital Twin}

The Digital Twin under consideration comprises three principal components:
Firstly, the real world, which in this project is limited to selected routes of the TAF-BW.
Data collection is primarily conducted at two significant junctions, utilizing a mobile sensor vehicle.
To ensure the privacy of public space recordings, traffic participant information is used in the form of time-tracked object lists.
These provide insights into the movement patterns of individual traffic participants, as well as facilitating conclusions regarding traffic density and user behaviour at intersections, for instance. Vehicular-to-everything (V2X) information is also recorded.
This technology is currently utilized by a limited number of vehicles and, in the context of the positional information it contains, is superfluous to the object lists recorded by the camera and LiDAR.
As part of the V2X information, the traffic light signal phases of the signaling systems are also recorded, so that specific phase states can be analyzed online in combination with traffic flow.
In order to reuse existing standards, the object lists and the signaling phases were sent frame by frame, with each traffic participant defined by pose, speed, identification, dimension and classification (see \cite{zipfl_traffic_2020}).

The second part describes the virtual side of the Digital Twin. The simulation engine is a standalone system that runs on its own and can process data from the real world and from various co-simulation interfaces. The engine provides the basis for the environment model with physics calculation and serves as a simulation master for synchronized control of various co-simulation modules. It is based on Unreal Engine 5.3 \cite{unrealengine5} and a custom framework developed to create and simulate traffic environments \footnote{A URL for the GitHub Repository will be provided for final submission}.

The third component of the Digital Twin is the interface for connecting to co-simulations and other interaction options with the simulation.
The co-simulation interfaces facilitate a heightened level of intricacy for discrete environmental components.
Although the simulation engine is capable of standalone operation, simulating traffic flows, vehicle driving behavior, trams and trains, and pedestrian movement, it is imperative that more sophisticated and thus more realistic behavior models can be simulated for a comprehensive analysis of road traffic.
This necessity extends to the calculation of traffic signal sequences.
In addition to purely digital simulations, a bicycle simulator has been developed to more accurately reflect the behavior of cyclists and facilitate more detailed investigation.
\begin{figure*}[ht!]
  \centering
  \def\svgwidth{0.9\textwidth}
  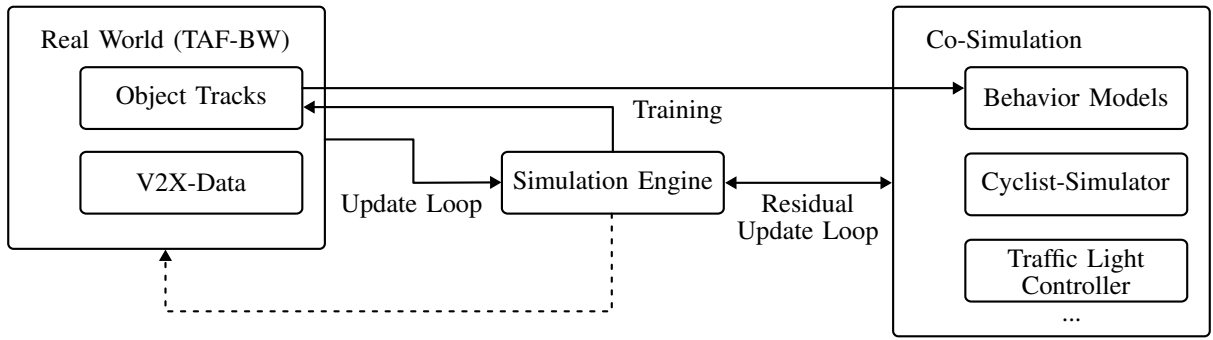
  \caption{Modules of the Digital Twin}
  \label{fig:strucutre}
\end{figure*}

\subsection{Object Detection and Tracking}

To reduce the gap between simulation and the real world, we integrate real-world human behavior models for dynamic traffic participants into the Digital Twin. To enable that, we use neural-network-based object detectors to detect traffic participants in TAF-BW. The detection methods are two-fold: Using cameras of smart infrastructures in two German cities of Karlsruhe and Heilbronn, complemented by additional detection capabilities provided through the mobile control center of TAF-BW.

Multiple traffic intersections in Karlsruhe and Heilbronn are equipped with pole-mounted cameras, strategically positioned to provide comprehensive coverage of intersection areas. To calibrate the cameras, satellite imagery is used, as in \cite{ramenau_2023}. Due to distortion and terrain, direct projection between camera and satellite planes is not feasible. Instead, a dynamic method adjusts the homography matrix per image location, using at least 30 manually selected pair of points. New points are projected using the seven nearest-labeled ones. Building upon our previous camera-based object detection approach~\cite{fleck2018, zipfl_traffic_2020}, we refine our detection capabilities through fine-tuning the object detectors with approximately 3,000 manually labeled images collected from various cameras across diverse traffic intersections. These images are captured at five-minute intervals over multiple days, thus enhancing the dataset's diversity.

Additionally, ~200,000 synthetic images were generated using the proposed Digital Twin and the CARLA~\cite{dosovitskiy2017carla} environment, allowing automatic generation of precise ground-truth labels for traffic participants. Synthetic data significantly increases dataset diversity by circumventing limitations posed by a finite number of real-world camera positions. Specifically, our dataset comprises eight real-world camera positions augmented by 800 synthetically generated camera positions.

\begin{figure}[h]
    \centering
    \includegraphics[width=1.0\linewidth]{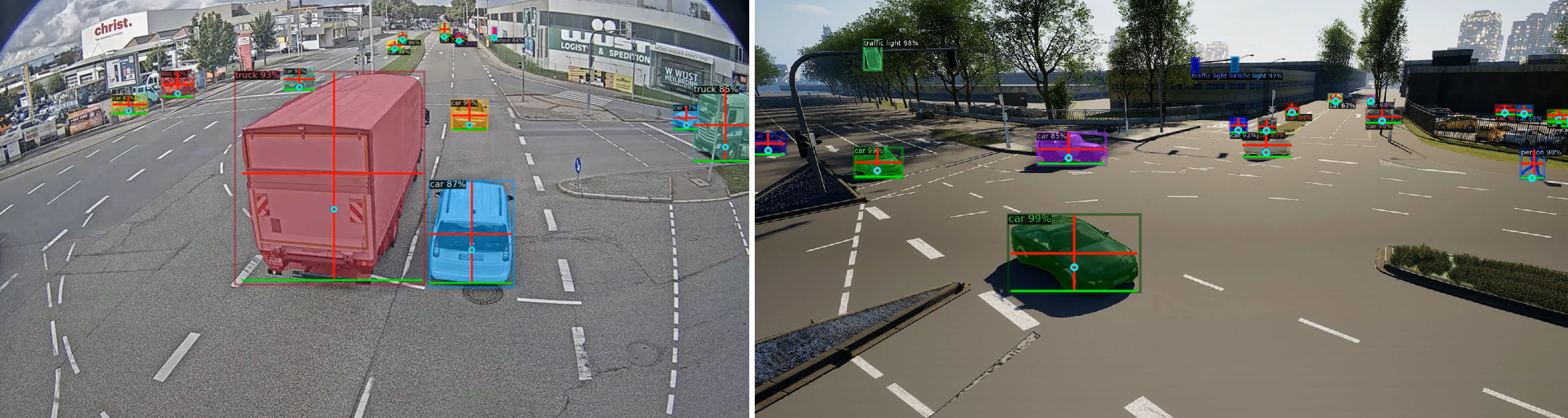}
    \caption{Traffic participants detection on real (left) and synthetic data (right) }
    \label{fig:realvssim}
    \vspace{-2ex}
\end{figure}

Traffic participants are detected via the Detectron2 \cite{wu2019detectron2} instance segmentation model, which produces object masks. Due to anonymization needs and camera limits, only position and class are extracted. The projection point is based on the base center of the mask, adjusted vertically to reduce distortion. Each camera processes data independently, and projected points are merged by proximity. Deviations are reconciled by computing a central joint point per group of objects. A tracker then links objects over time by minimizing distance between predicted and new positions. Finally, tracks are converted to geocoordinates as described in \cite{ramenau_2023} and formatted to match the TAF-BW dataset \cite{zipfl_traffic_2020}, including ID, timestamp, type, position, rotation and speed.
\Cref{fig:realvssim} shows examples of synthetic and real images and applied object recognition.

TAF-BW's mobile control centre is a large van equipped with a sensor suite of multiple LiDARs and cameras to capture the environment. In total, there are six LiDARs, one mid-range sensor in each corner of the roof and another one on the driver's side, as well as one long-range sensor located centrally in the front. This results in a full 360° surround view using the LiDARs only. In addition, stereo vision is enabled through two cameras mounted on top at the front. A high accuracy coupled GNSS+INS with two antennas provides precise localization. 
The data is processed and stored by a high-performance computer with two GPUs.

The creation of the object lists is based on a LiDAR-based 3D object detection model using the OpenPCDet framework \cite{openpcdet2020}. The used models are all pre-trained on the KITTI-dataset \cite{Geiger2012CVPR}. Through a given benchmark by OpenPCDet, which is based on KITTI data, a pre-selection was performed. 
The system's primary detection focus is on dynamic traffic participants such as cyclists, pedestrians, and cars.
The objects are detected on individual, time-independent frames in static point clouds. The SimpleTrack algorithm \cite{pang2021simpletrack} is used to merge identical objects consistently over time in all frames.
To improve the performance of the 3D object detection, additional point cloud data was recorded in the area of the TAF-BW and manually annotated. This dataset, combined with data from an open-source dataset of MAN \cite{truckscenes2024}, was used to retrain the detection models. In total, the dataset contains 6.159 data samples. 
Out of the recorded point cloud data, the object list is created online as a ROS topic. Additionally, the object list is also saved as a csv-file. In total, 35 parameters are included, such as the position of the detected object, the yaw rate, or the relative velocity. The structure of the object list is aligned with the project's layout and therefore, with the recorded object lists at the intersections (camera recording). As the other object lists, this is the basis for the co-simulation.

\begin{figure}[h]
    \centering
    \includegraphics[width=\linewidth, trim=0 0.5cm 0 4cm, clip]{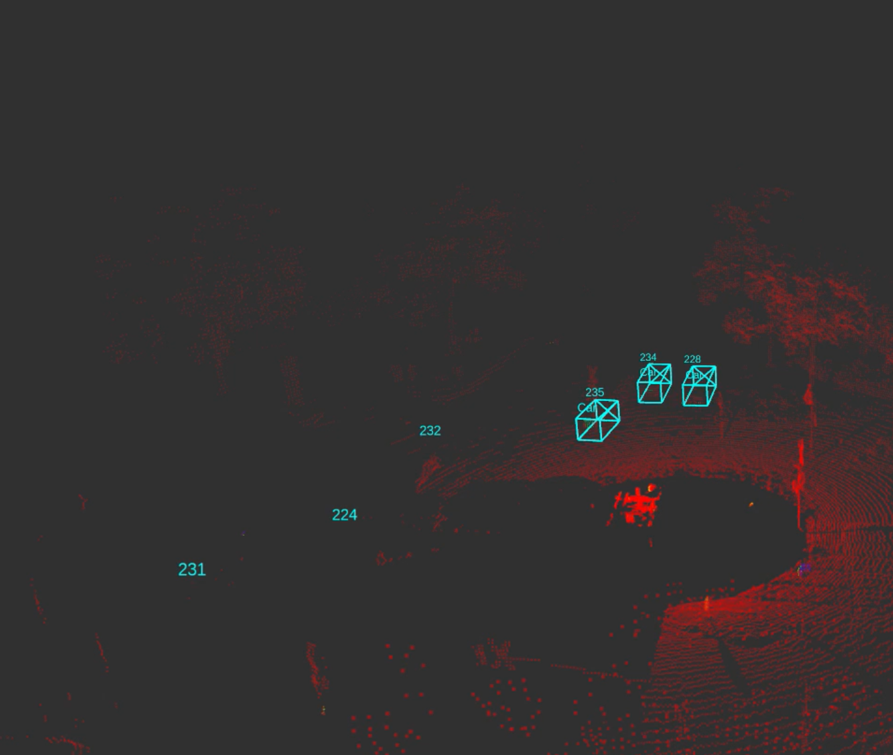}
    \caption{Example of a LiDAR point cloud (red dots) recorded alongside object detection (white wire boxes) of a traffic scene in TAF-BW}
    \label{fig:mobiler_leitstand_pointcloud}
\end{figure}

\Cref{fig:mobiler_leitstand_pointcloud} shows an example of the data recordings with the mobile control center.

\subsection{Creation of a 3D-World}

The digital environment should include 3.6\,km of urban roads, 3.1\,km of dual lane highways and over 600\,m of tunnels. For such a large scene, compared to Neural Radiance Fields (NeRFs) \cite{mildenhall2020nerf, huang2023refsr, tang2024parenerf} or Gaussian Splatting \cite{tang2023efficient, wang2024improving, yao2025reflective}, only a mesh-based representation is suitable. It provides a dynamic scene that can be controlled by the user,  compatible with industry standards, and small enough in file size for public distribution. Modeling the complete road network by hand would require a significant amount of time, so we use EG4U \cite{schulz2023fast, eg4u2025github} for generating the background city automatically, combined with a custom-developed framework that enables us to rapidly place and generate roads, tram tracks, pedestrian pavement, and generic environment elements. This hybrid approach of creating the virtual world allows us to build a detailed Digital Twin. Additionally, we use aerial, satellite, photogrammetry, and OpenStreetMap data \cite{opengeodata2025} to model the environment according to real-world locations. The modeling is mainly done using our framework, which provides the following features:
\begin{itemize}
    \item Street and environment editor directly in Unreal Engine
    \item Procedural generation of dense foliage such as grass, trees, and bushes
    \item Integration into Unreal Engine's Blueprint system, extensible through scriptable modeling capabilities 
    \item Highly efficient traffic simulation logic directly integrated into the framework modeling tools
\end{itemize}

The following is a brief explanation of how the 3D modeling of a real-world area is carried out in Unreal Engine.
In order to obtain a general overview of the area and, at the same time, to be able to reproduce important landmarks such as buildings and roads, low-resolution objects are loaded into Unreal using the OSM tool.
It is important to note that all georeferenced maps, such as OSM and aerial footage, are mapped to a flat 2D plane using the Mercator projection.
The objects created with data from OSM are inadequate for our use case, which is to create an accurate and photorealistic replica of the real world.
Consequently, we also employed aerial photographs \cite{opengeodata2025} and digital surface models of the area to ascertain the precise positions of individual objects and streets.
For instances where no such information is available, satellite imagery can be used, although at a lower resolution.
Both aerial and satellite imagery must be georeferenced so that they can be matched with the OSM data. To account for different projection spaces of OSM and aerial images, we rely on a custom script using rasterio \cite{Gillies2013rasterio} to transform both into Mercator.
Using the aerial images as a ground plane, the OSM objects are placed on top of them in the simulation engine.

The subsequent stage of the process is to generate the streets and directly with them associated elements.
The following description of an elementary road section is based on the OpenStreetMap definition.
It is important to note that each section of a road's cross section is defined individually.
This road segment can contain several lanes, grass verges, or lane markings.

\begin{figure}
    \centering
    \includegraphics[width=0.45\linewidth]{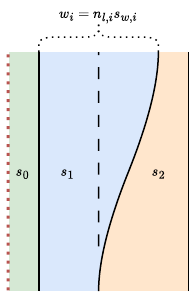}
    \includegraphics[width=0.49\linewidth]{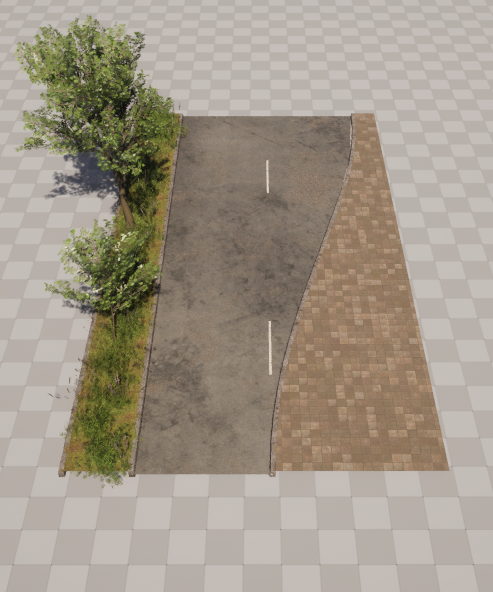}
    \caption{An example schematic defining a street segment. The width of the individual segment parts depends on the number of lanes $n_l$ and the width of a single segment lane $s_w$. On the right, the resulting segment in Unreal Engine is shown.}
    \label{fig:sim_segments}
\end{figure}

In \Cref{fig:sim_segments}, we show an example street segment containing one vegetation track $s_0$ followed by a road segment $s_1$ and ending with a pedestrian path $s_2$. The different parts $s_{0,1,2}$ are defined as blueprint classes and contain parameters for the default width, materials, and code for placing assets and foliage. Apart from the width $w_i$ of a segment part $i$, we also use a height offset for the surface. This allows us to place vegetation and pedestrian pavement above the road level split by a curb, as seen in the right image in  \Cref{fig:sim_segments}.

To enable the generation of every theoretical road layout, each individual road element and its associated sections can be freely parameterized.
The present implementation necessitates manual input of the parameterization, though this can be automated to a considerable extent by assuming typical distributions of parameter sets that occur in a specific section of the area to be modeled (for example, garden fences, green strips, trees) or incorporating OSM data as a baseline. The asset placements procedural generator supports three different asset sampling strategies: random, round-robin, and regular expressions (regex), which are required when there are multiple variations of a mesh. The regex sampler allows for adding constraints to the generation process; for example, when generating a fence or house wall, we only want one gate or door to be present. Currently, it supports the following operations:

\begin{itemize}
    \item $A*$- Randomly sample as many assets as possible from set $A$
    \item $A+$ - Randomly sample at least one asset and as many as possible from set $A$
    \item $A?$ - Randomly sample zero or one asset from set $A$
    \item $(A|B)$ - Randomly sample one asset from set $A$ or $B$
\end{itemize}

Concatenations, such as $(A|B)*CA+$, are also possible.

\begin{figure}
    \centering
    \includegraphics[width=1.0\linewidth]{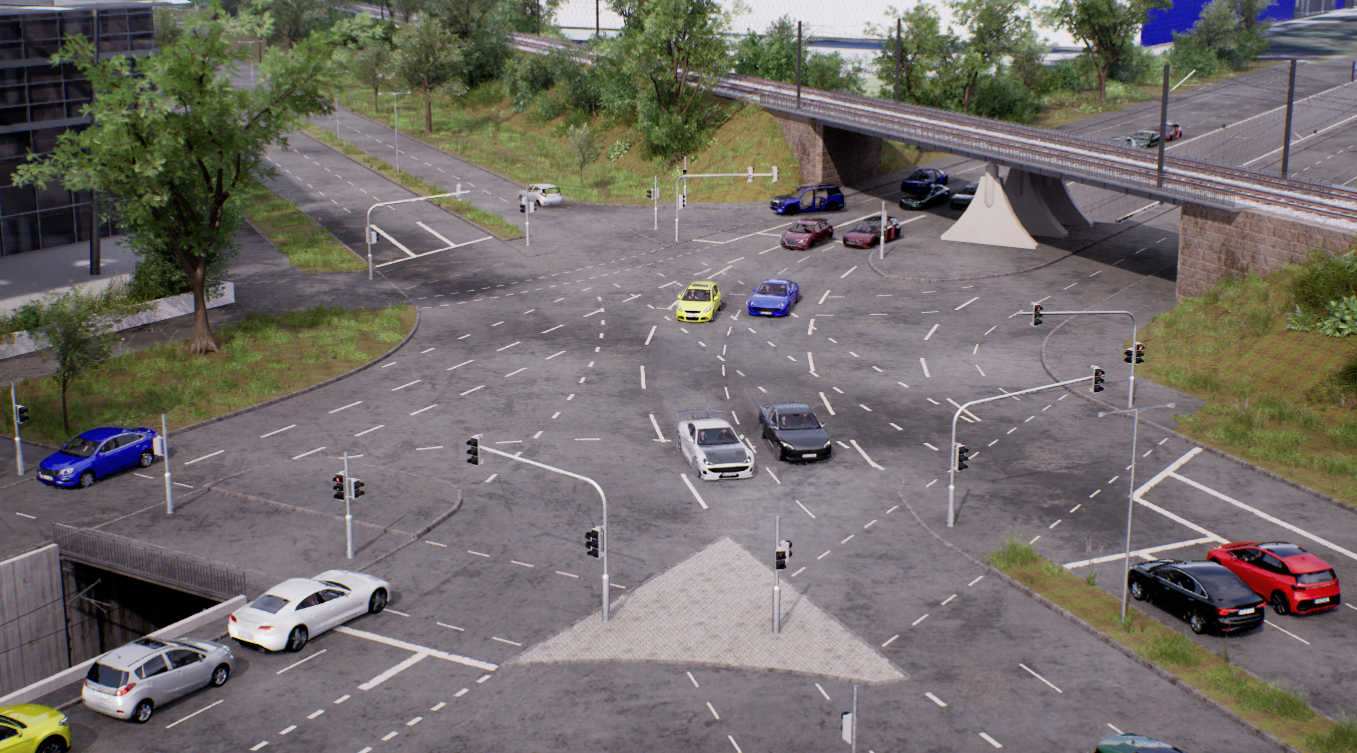}
    \caption{How a junction looks in our simulation, specifically the K733 connecting the Ostring and Durlacher Allee in Karlsruhe, Germany.}
    \label{fig:sim_junction}
\end{figure}

Subsequent to the modeling of linear streets, intersection objects are incorporated to establish links between intersecting streets or converging streets and their street topology. The base surface model, as well as the lane connections, are automatically generated and can be adjusted by the user to match the underlying road network. In  \Cref{fig:sim_junction}, the intersection with the id "K733" of our environment model is shown. Lane markings, traffic lights, and traffic islands are manually placed using our framework. The intersection itself contains a controller for traffic lights, which automatically connects the manually placed ones with the different signal groups. Also, pedestrian crossings contain navigation data to indicate a red or green light.

The last elements needed for the environment are buildings, foliage, and generic objects. For this purpose, our framework can communicate basic shape data between streets, splines, buildings, etc., that support the user to fill out gaps in the environment, e.g., foliage between the street and the railway in \Cref{fig:sim_junction}. This interoperability saves time compared to manually placing a mesh at the exact position to make a watertight surface. Watertight in this scenario means that there is not a single pixel gap at any edge where the camera could see through the surface.

\subsection{Co-Simulation and Behavior Models}
For a complex interplay between different simulations and complex behavior models, connecting other simulators and models via TCP or UDP is a key feature of the Digital Twin. To exchange data, we use a format similar to the one used in previous works \cite{zipfl_traffic_2020, fleck2018}. During co-simulation, the environment simulation broadcasts its complete internal state for the current simulation frame, including positions and attributes of traffic participants and infrastructure objects. All connected models or simulators can then push their updated object information back to the environment simulation, where all the data is aggregated and published again as the new simulation state. The environment simulation also incorporates tools for recording and playing back traffic scenarios. This scenario player uses the same data structure and can also be used for over-dubbing recorded scenarios, e.g., recording different traffic participants after one another. When re-simulating scenarios, one can change environment parameters, such as the time of day, weather conditions, or the current season, which ultimately affects the scenery. The most notable effects include foliage and road conditions, where increased precipitation leads to wet surfaces or puddles and, in combination with different seasons, fallen leaves or snow on the ground \cite{schulz2024}. This is particularly interesting for creating variation-rich datasets.

In order to incorporate dynamic objects into the simulation, with a particular focus on vehicles and road and railway-bound agents, a basic behavior model is implemented. This model can be applied to any dynamic object without the need to rely on external behavior models.
The environment is populated with vehicles and trains spawned at the beginning of a lane or train track.
A vehicle $i$ gets a random velocity $v_\mathrm{set} = v_\mu + \xi_i v_\sigma$ for a mean velocity $v_\mu$ with a variation $v_\sigma$ and a random variable $\xi_i \in [-1, 1]$. For dynamic driving behavior, a vehicle has a max. acceleration $a$ and a max. breaking acceleration $a_b$ in the unit $[\textrm{m}/\textrm{s}^2]$. An Euler integrator is used for calculating the vehicle dynamics.

To react to the environment, every vehicle is controlled by a virtual driver. The driver is only told to stop at a certain distance $d_\mathrm{stop}$ from the current position if there is any obstacle, like a person, car, tram, etc. in the way of driving. Using $d_\mathrm{stop}$, we calculate a target velocity $v_\mathrm{target}$ for the vehicle the following way:

\begin{align}
    a_\mathrm{max} &= \frac{3 a_b}{4}\\
    t_0 &= \frac{v_\mathrm{set}}{a_\mathrm{max}}
\end{align}
\begin{align}
    dx &= v_\mathrm{set} t_0 - \frac{1}{2} a_\mathrm{max} t_0^2\\
    v_\mathrm{target} &= \mathrm{lerp(v_\mathrm{obs} \frac{d_\mathrm{stop}}{50}, v_\mathrm{set}, \frac{d_\mathrm{stop}}{dx})}
\end{align}
with $v_\mathrm{obs}$ being the velocity of the obstacle relative to the vehicle. The "pedal" value controlling the acceleration and breaking of the vehicle is then given by $\mathrm{pedal} = clamp(v_\mathrm{target} - v_\mathrm{cur}, -1, 1)$. $lerp(a,b,t)$ is defined as the linear interpolation between $a$ and $b$ using $t$ and $clamp(t,a,b)$ clips the value $t \in [a,b]$.

To model pedestrian behavior realistically, Unreal Engine is used in combination with MetaHuman \cite{metahuman} for high-fidelity digital humans. Skeletal meshes define bone structures and enable the animation of head and body movements. Pedestrian trajectories follow splines with continuously updated positions and rotations, simulating natural locomotion. Behavioral parameters such as looped paths, stopping behavior, and time constraints are configurable. Head orientation is controlled via an animation blueprint, a look-at mesh (focus), and collision detection. When the focus enters the pedestrian’s detection range, the head rotates smoothly toward it using interpolated transitions, synchronized with walking or running animations depending on speed.

\begin{figure}[h]
    \centering
    \includegraphics[width=1.0\linewidth]{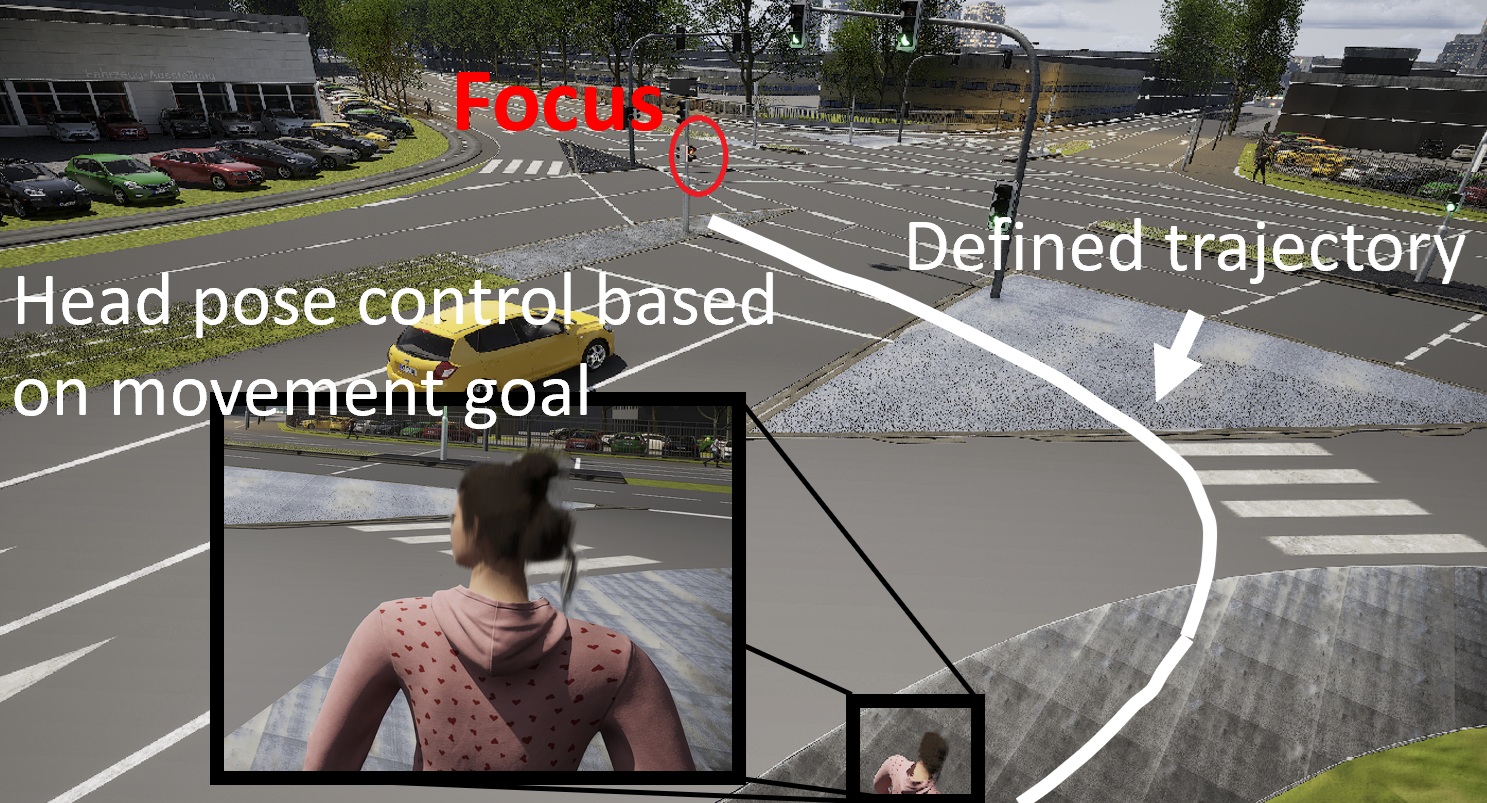}
    \caption{Pedestrian behavior model with controlled head pose, goal focus, and defined trajectory}
    \label{fig:pedsim}
\end{figure}

\begin{figure}
    \centering
    \includegraphics[width=1.0\linewidth]{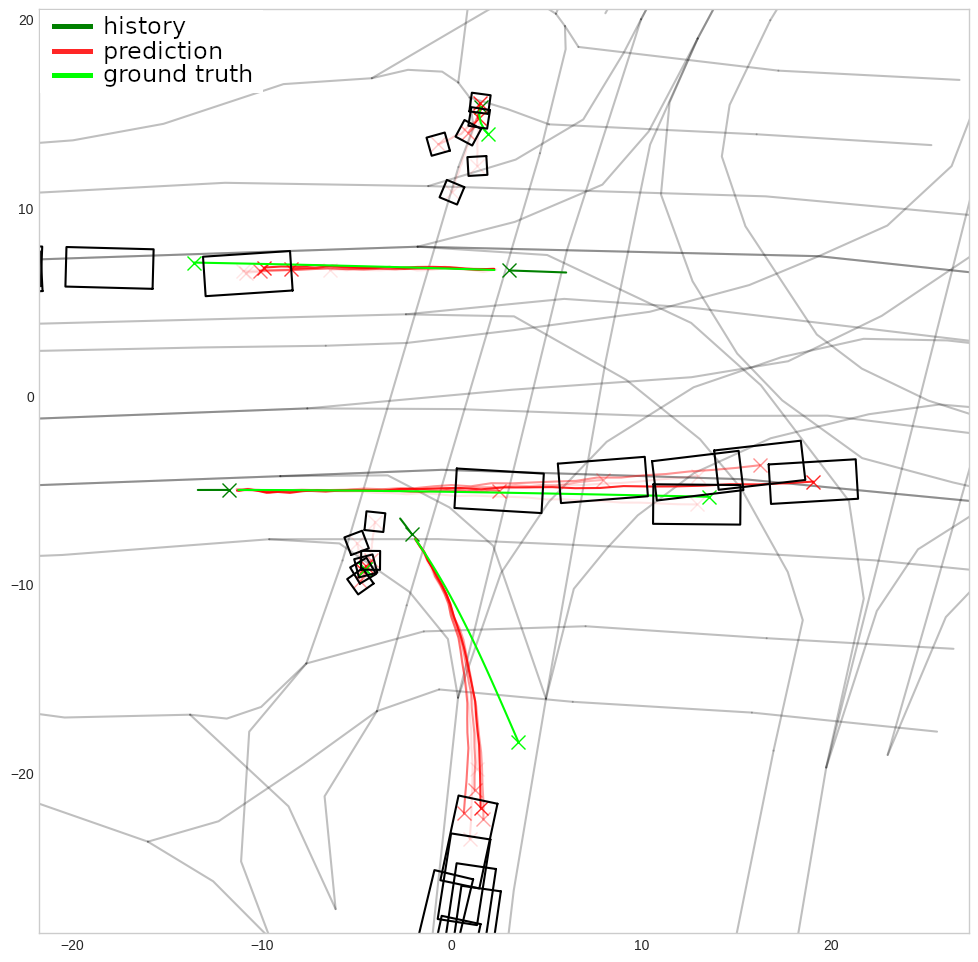}
    \caption{Simulated multi-agent future trajectories using our proposed behavior model.}
    \label{fig:holigraph_prediction}
\end{figure}

Besides the aforementioned rule-based model, we additionally provide a more complex, data-driven behavior predictor to support use cases that require deviation from ordinary, expected behavior. As an example, our Digital Twin might be applied for scenario-based testing for automated vehicles, where traffic participants of a scene ought to react even to particularly rare edge-case situations in a realistic way, that might not be captured by a purely analytical model. To this end, we build upon "HoliGraph", the work of Grimm et al. \cite{grimm_heterogeneous_2023} and employ a graph neural network-based encoder-decoder architecture to perform multi-agent trajectory prediction. Given a variable-length sequence of past scenes encoded as a heterogeneous, spatio-temporal graph, the model extrapolates them to the future, thus acting as a closed-loop simulation. Using a pre-trained (on the nuScenes dataset\footnote{\url{https://www.nuscenes.org/}}) variant of the model to simulate scenarios obtained in TAF-BW yields \textit{minADE} and \textit{minFDE} scores of 0.325 and 0.656, respectively, for the most likely future ego trajectory. \Cref{fig:holigraph_prediction} depicts a set of predicted futures of an exemplary \text{TAF-BW} scene. Ultimately, the behavior model -- running either online or offline -- can be used to explore the anticipated effects of changes in the environment, such as altered road layout, signal phases or traffic rules.

\begin{figure}
    \centering
    \includegraphics[width=1.0\linewidth]{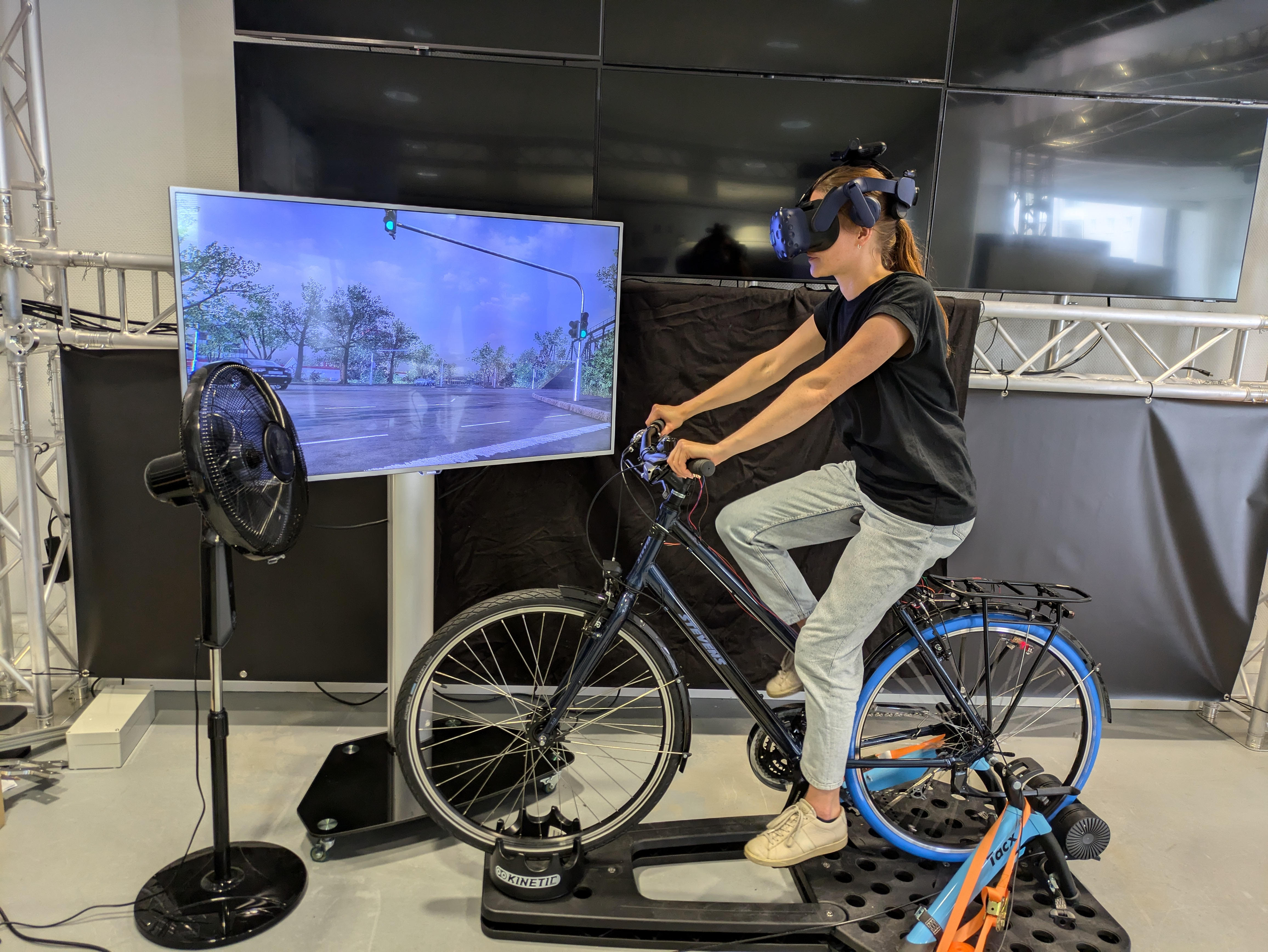}
    \caption{Co-Simulation with our bicycle simulator.}
    \label{fig:bicycle_simulator}
\end{figure}

The Digital Twin also offers the possibility of co-simulation with hardware simulators. As an example, we have built and integrated a bicycle simulator shown in \Cref{fig:bicycle_simulator}. The simulator allows users to engage in realistic cycling scenarios through the Digital Twin, such as varying weather conditions and different traffic volumes, without the need for outdoor cycling. The virtual representation of the actual bike adapts to the speed and steering angle of the bike. The data is measured by sensors attached to the bike and sent via Bluetooth Low Energy to the computer on which the simulation is running and through TCP to the Digital Twin. This synchronization allows for a more accurate and responsive simulation, where the behavior of the bicycle in the virtual environment mirrors that of the physical bike, creating a seamless user experience. This allows cyclists' behavior to be studied and data to be recorded safely \cite{gremm_ICSC}.

% =============
%  Experiments
% =============
%%%%%%%%%%%%%%%%%%%%%%%%%%%%%%%%%%%%%%%%%%
\section{Experiments}
\label{sec:experiments}
%%%%%%%%%%%%%%%%%%%%%%%%%%%%%%%%%%%%%%%%%%

\subsection{Traffic Light Optimization}
Traffic signals influence both traffic participant behavior and overall traffic quality. Modifications to traffic-responsive signal control can be integrated into a Digital Twin to evaluate effects relative to baseline conditions. For instance, optimized signal control strategies for pedestrians and cyclists were simulated to minimize lost time.
The TAF-BW traffic detection system provides previously unavailable data, including participant detection and trajectory information. If traffic participants could transmit individual signal control requirements (e.g., destination after passing the intersection), further data would enhance signal responsiveness and individualization.
The combination of detailed trajectory data and potential V2X-based cooperation between participants and infrastructure increases both the complexity of signal control logic and planning demands.
This offers potential by supporting decision-making through real-time analysis of participant movement patterns near the intersection.

To demonstrate how the integration of this additional data would impact pedestrians and cyclists (VRU), achievable optimization potential was determined using a simulation of a large, four-arm intersection. Recorded traffic data was used for this purpose, and simulations were repeated several times with various modifications.

The infrastructure, the traffic-dependent traffic control of the traffic signal system, and the traffic volume at the intersection Karl-Wüst-Straße/Austraße in Heilbronn, Germany were mapped in a simulation environment.  Simulations were conducted over one hour each under real traffic conditions at four different times of day, using different traffic flows and traffic control programs.  In addition, pedestrian an bicycle traffic volumes were increased in some simulations compared to reality. Subsequently, the simulations were repeated with various optimization measures in the traffic signal control system, and the time losses for traffic participants were recorded. The optimization potential was derived from the comparison of the optimizations (opt.) with the existing situation without optimizations (n.opt.). The optimizations to the signal control system essentially consisted of allowing pedestrians to cross the main direction via two consecutive crossings without stopping on the central island and ensuring that the green time were caught. On the other hand, motor vehicle traffic could be granted more green time when there were no more pedestrians or cyclists.

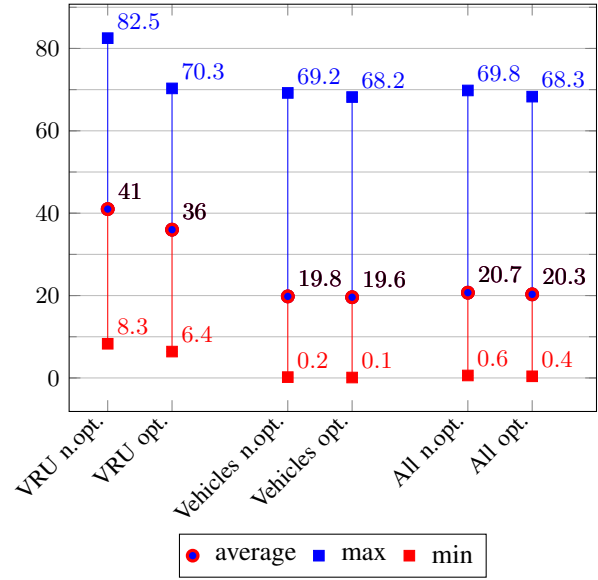
\begin{figure}[h]
    \centering
    \pgfplotstableread{
x y y-max y-min
0     41.0 41.5 32.7
0.5   36.0 34.3 29.6
1.4   19.8 49.4 19.6
1.9   19.6 48.6 19.5
2.8   20.7 49.1 20.1
3.3   20.3 48.0 19.9
}{\differanser}

\begin{tikzpicture}
\begin{axis} [
    width  = 0.99\columnwidth,
    height = 7cm,
    xtick={0,0.5,1.4,1.9,2.8,3.3},
    xticklabels={{VRU n.opt.},{VRU opt.},{Vehicles n.opt.},{Vehicles opt.},{All n.opt.},{All opt.}},
    minor ytick={0,10,20,30,40,50,60,70,80,90},
    yminorgrids,
    ticklabel style = {font=\small},
    x tick label style={rotate=45,anchor=east},
    xmin = -0.3,
    xmax = 3.8,
    every node near coord/.append style={font=\small, anchor=south west},
    nodes near coords,
    point meta=y,
    legend style={
        at={(0.5,-0.3)},
        anchor=north,
        cells={anchor=west},
        column sep=1ex
    },
    legend columns=-1
]
\addplot+[blue, very thick, forget plot, only marks] 
plot[very thick, error bars/.cd, y dir=plus, y explicit]
table[x=x,y=y,y error expr=\thisrow{y-max}] {\differanser};

\addplot+[red, very thick, forget plot, only marks] 
plot[very thick, error bars/.cd, y dir=minus, y explicit]
table[x=x,y=y,y error expr=\thisrow{y-min}] {\differanser};

\addplot[only marks,mark=*,mark options={fill=blue,draw=red,very thick}] 
table[x=x,y=y] {\differanser};
\addlegendentry{average}

\addplot[only marks,mark=square*,color=blue] 
table[x=x,y expr=\thisrow{y}+\thisrow{y-max}] {\differanser};
\addlegendentry{max}

\addplot[only marks,mark=square*,color=red] 
table[x=x,y expr=\thisrow{y}-\thisrow{y-min}] {\differanser};
\addlegendentry{min}
\end{axis} 
\end{tikzpicture}
 \caption{Time Lost -  Example of results from the traffic signal analysis (time period 4:00 p.m. - 5:00 p.m., 3,287 vehicles, 135 VRUs)}
    \label{fig:AvgTimeLost}
\end{figure}

With the optimizations in the traffic signal control system examined, the average and maximum time losses for pedestrians and cyclists could be significantly reduced without increasing the time losses for motor vehicle traffic. Even with higher VRU volumes, the results for motor vehicle traffic remained stable. The optimizations reduced the average time losses for pedestrians and cyclists by approximately 10-20\% in all time periods examined, even with higher volumes of VRUs. \Cref{fig:AvgTimeLost} shows the results for the afternoon peak hour in the variant in which the number of VRUs was increased. In the afternoon, the intersection experienced the highest traffic volume, with 3,287 vehicles, and thus the most unfavorable conditions for pedestrians and cyclists. The reduction in the maximum time losses contributes to a better assessment of the intersection's traffic quality. In terms of traffic safety, it can be assumed that red light violations by pedestrians and cyclists will decrease if the high individual waiting times for pedestrians and cyclists are reduced, thereby reducing the risk of an accident.

\subsection{Security Analysis}
The close integration between V2X measurement data and observed object tracks allows for novel approaches in security analysis of vehicular communication.
As most of the security requirements relate to the impact on the physical world, our Digital Twin helps to understand critical scenarios and relevant mitigation techniques.
We applied our framework by first developing a threat model for connected and automated driving.

We applied a two-step process used for establishing and scoring threats loosely based on the ISO 21434 standard with adaptions to fit our generalized architecture for connected vehicles.
The resulting 36 threats were scored according to the estimated likelihood of exploitation multiplied with the maximum damage potential in the domain's traffic efficiency, safety, privacy and authenticity.
As a result, we assigned the highest scores to the following threats:
\begin{itemize}
    \item Loss of privacy
    \item Traffic flow manipulation with spoofed message contents
    \item Continued broadcasting of malicious messages because of deferred revocation
    \item Injecting false data on internal vehicle busses
    \item Manipulation of sensor readings
\end{itemize}

\begin{figure}[h]
    \centering
    \includegraphics[width=1\linewidth]{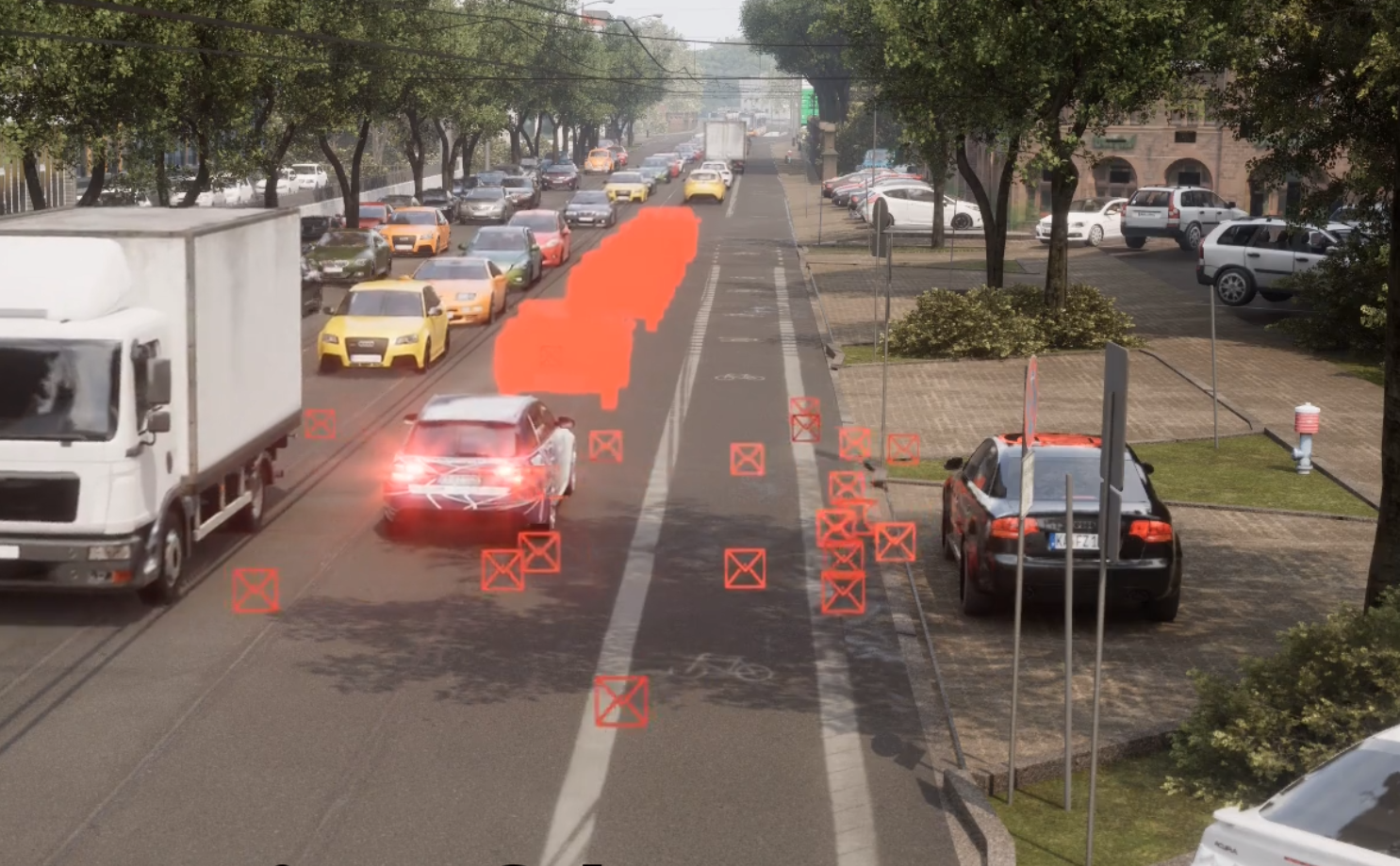}
 % \vspace{-0.4cm}
 \caption{Photorealistic false data injection attack visualization showing the spoofed vehicles as a red overlay and the effect of the victim decelerating to avoid a potential collision.}
    \label{fig:attack-vis}
\end{figure}

We used the capabilities of the Digital Twin to demonstrate the plausibility of a data spoofing attack shown in Figure \ref{fig:attack-vis}.
Our workflow facilitates the creation and demonstration of highly realistic attack scenarios that are otherwise difficult to obtain and can be used to develop and evaluate effective misbehavior detection systems.

% ============
%  Conclusion
% ============
%%%%%%%%%%%%%%%%%%%%%%%%%%%%%%%%%%%%%%%%%%
\section{Conclusion}
\label{sec:conclusion}
%%%%%%%%%%%%%%%%%%%%%%%%%%%%%%%%%%%%%%%%%%

In this work, we presented a comprehensive Digital Twin framework for smart urban mobility based on the TAF-BW infrastructure. Our system bridges the physical and digital domains through detailed modeling, real-time data integration, and co-simulation capabilities, enabling scenario-based analysis and optimization. By focusing on traffic signal optimization and V2X security, we demonstrated the practical utility of the DT in enhancing both efficiency and safety. The integration of behavior models, including vulnerable road user (VRU) simulations, highlights the system's flexibility. Future work will focus on improving behavioral realism and conducting long-term evaluations in different urban and smart-city contexts. The publicly available tools aim to support further research and development in mobility digital twins.

\section{Acknowledgements}
The work leading to these results is part of the service contract “AI in mobility” on behalf of the Ministry of Transport Baden-Württemberg.

\bibliographystyle{IEEEtran}
\bibliography{bib_options,references}
\end{document}